\documentclass{article} 
\usepackage{iclr2026_conference,times}

\usepackage{amsmath,amsfonts,bm}

\def\eqref#1{equation~\ref{#1}}

\def\1{\bm{1}}

\DeclareMathAlphabet{\mathsfit}{\encodingdefault}{\sfdefault}{m}{sl}
\SetMathAlphabet{\mathsfit}{bold}{\encodingdefault}{\sfdefault}{bx}{n}

\usepackage{hyperref}
\usepackage{url}

\usepackage{enumitem}
\usepackage{graphicx}
\usepackage{booktabs}
\usepackage{algorithm}
\usepackage{algpseudocode}
\usepackage{amsmath}
\usepackage{amssymb}
\usepackage{xcolor}
\usepackage{xspace}
\usepackage{subcaption}
\usepackage[table]{xcolor}
\newcommand{\udvlarl}{\textbf{dVLA-RL}\xspace}
\title{dVLA-RL: Reinforcement Learning over Denoising Trajectories for Discrete Diffusion Vision-Language-Action Models}

\author{
  \vspace{-25pt}\\
  \textbf{Yuhao Wu$^{1,2*}$,\ \ Yitian Liu$^{1,3*}$,\ \ Weijie Shen$^{1,4}$\thanks{Co-first authors, equal contribution. $^{\dagger}$Corresponding authors. $^{\ddagger}$Project co-leaders. This work was completed by Yuhao Wu during his internship at ScaleLab at SJTU. Wei Sui at D-Robotics and Ru Ying at Baidu AI Cloud are co-leaders of this project.}\ \ ,\ \ Mishuo Han$^{1}$,\ \ Wenjie Xu$^{3}$,\ \ Haotian Liang$^{5}$}\vspace{3pt}\\ 
  \hspace{3pt}\textbf{Zhongshan Liu$^{3}$,\ \ Yinan Mao$^{3}$,\ \ Lei Xu$^{1}$,\ \ Xinping Guan$^{1}$,\ \ Ru Ying$^{3\ddagger}$,\ \ Ran Zheng$^{3}$}\vspace{3pt}\\
  \hspace{3pt}\textbf{Wei Sui$^{4\ddagger}$,\ \ Xiaokang Yang$^{1}$,\ \ Wenbo Ding$^{2}$,\ \ Yao Mu$^{1,5 \dagger}$}\vspace{3pt}\\
  $^1$Shanghai Jiao Tong University\vspace{1pt}
\\$^2$Tsinghua Shenzhen International Graduate School, Tsinghua University\vspace{1pt}
\\$^3$Baidu AI Cloud \quad\quad $^4$D-Robotics\vspace{1pt} \quad\quad $^5$Shanghai AI Laboratory
}

\iclrfinalcopy 
\begin{document}

\maketitle
\thispagestyle{empty}
\pagestyle{empty}
\begin{abstract}
Vision-Language-Action (VLA) models have established a powerful paradigm for generalist robotic manipulation by grounding control into the semantic reasoning of large Vision-Language Models (VLMs). Prevailing architectures typically model actions continuously via diffusion or flow processes, or discretely through either autoregressive generation or parallel decoding. Recently, Discrete Diffusion VLAs (dVLAs) have emerged as a distinct alternative, unifying vision, language, and action into a single discrete token space via masked generative modeling. While this paradigm successfully combines iterative action refinement with unified representations, its training has thus far been restricted to Supervised Fine-Tuning (SFT), leaving the potential of Reinforcement Learning (RL) for further policy refinement largely unexplored. A fundamental challenge in RL for dVLAs is that the marginal probability of the final action generated by dVLAs remains intractable. To solve this problem, we propose \udvlarl, shifting the learning objective from the marginal action probability to the joint probability of the sampled generation path. Specifically, by modeling the denoising process as a Markov Decision Process (MDP), we mathematically formulate this path probability as a product of step-wise transitions. This trajectory-level objective provides a unified formulation that natively accommodates variable denoising steps. Leveraging this intrinsic flexibility, we introduce a unified step scheduling approach for complex multi-task learning, tailoring denoising steps to specific task complexities to maximize both success rates and computational efficiency. Extensive evaluations demonstrate that our approach achieves a success rate of \textbf{99.7\%} on LIBERO. Furthermore, it establishes strong VLA-based results on RoboTwin 2.0 by delivering a \textbf{30.6\%} improvement over the SFT baseline, remaining competitive with strong World-Action Model baselines.
\end{abstract}

\begin{figure}[ht]
    \centering
    \includegraphics[width=\linewidth]{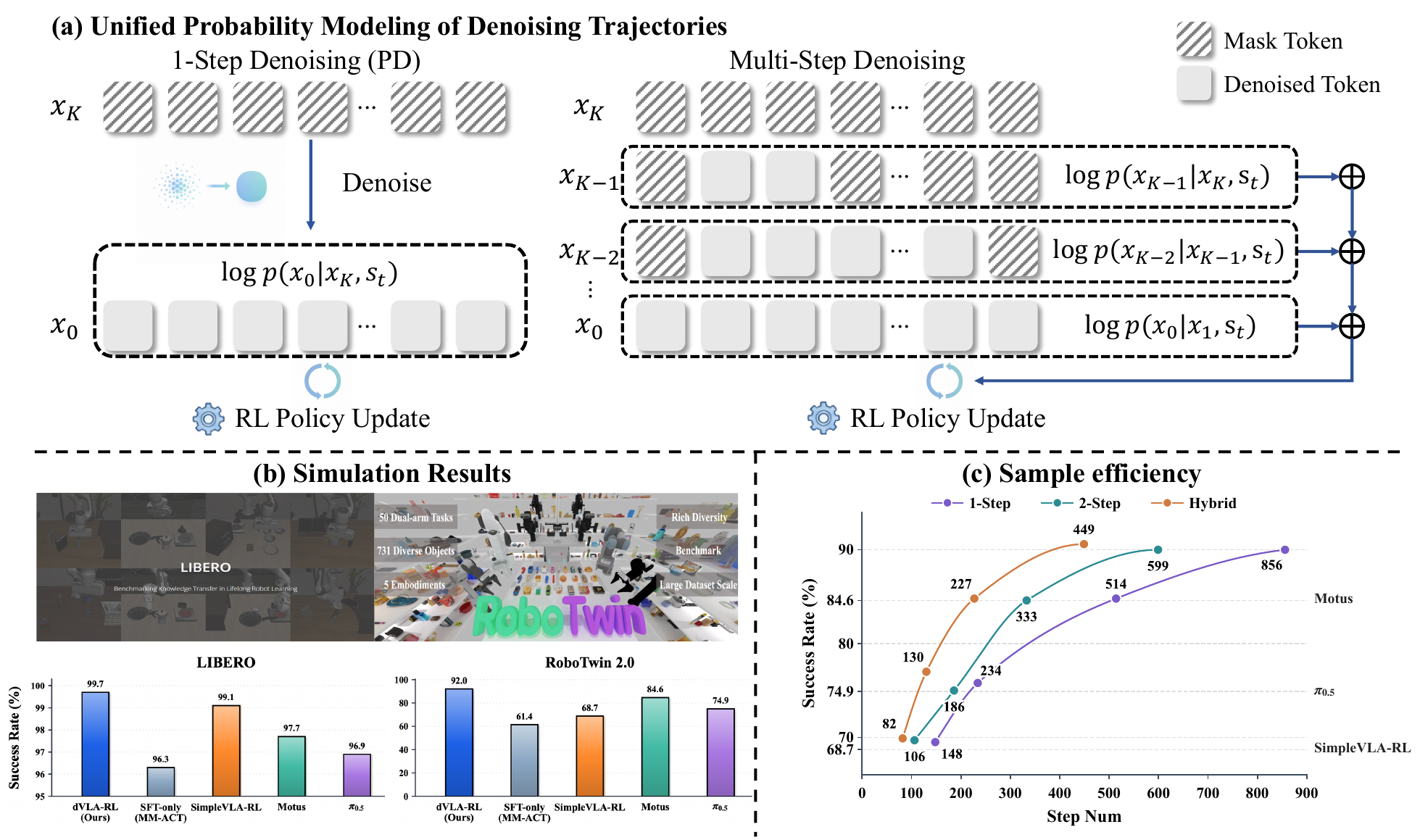} 
    \caption{\textbf{Overview of UDVLA-RL.} \textbf{(a)} We present a unified probability modeling approach for denoising trajectories, seamlessly integrating 1-step parallel decoding and multi-step denoising paradigms. \textbf{(b)} Extensive evaluations on the LIBERO and RoboTwin 2.0 benchmarks demonstrate that our RL-finetuned policy outperforms both SFT-only baselines and remains highly competitive among modern VLA architectures, achieving performance comparable to World-Action Models. \textbf{(c)} Our hybrid version exhibits superior sample efficiency, achieving higher success rates with substantially fewer environment interaction steps.}
    \label{fig:teaser}
\end{figure}

\section{Introduction}
\label{sec:intro}

Grounding multimodal perception into precise physical actions remains a central challenge in Embodied AI. Vision-Language-Action (VLA) models have advanced robotic control by framing it as a sequence generation task~\citep{pi_0, intelligence2025pi05visionlanguageactionmodelopenworld, openvla, openvla-oft}. However, their reliance on Supervised Fine-Tuning (SFT) exposes them to distributional shift, leading to compounding errors during real-world execution. Reinforcement Learning (RL) provides a principled solution to this bottleneck. By shifting the optimization target from mimicking static datasets to maximizing environment rewards, RL directly targets task success, unlocking unprecedented capability upper bounds for VLA policies~\citep{li2026simplevlarl}. However, applying policy-gradient RL to VLA policies requires a tractable policy likelihood for the generated action, making the underlying action-generation paradigm a crucial design factor.

Existing VLA action-generation paradigms employ distinct RL optimization strategies tailored to their underlying probability representations. Autoregressive and parallel decoding policies formulate action generation through token-wise or single-step predictions, allowing standard policy-gradient algorithms to directly utilize their exact analytical action likelihoods. Conversely, continuous diffusion and flow-matching policies focus on the expressive modeling of complex, multimodal action distributions. To facilitate RL in these continuous spaces, current methods typically evaluate policy gradients through likelihood approximations, variational objectives, or auxiliary stochastic formulations~\citep{park2025flowqlearning, liu2025flowgrpotrainingflowmatching, chen2026pitextttrlonlinerlfinetuning}.

Recently, Discrete Diffusion VLAs (dVLAs) have emerged as a promising new paradigm, unifying multimodal inputs and action chunks into a single discrete token space via masked generative modeling~\citep{liang2025discretediffusionvlabringing}. While recent works demonstrate strong Supervised Fine-Tuning (SFT) performance comparable to continuous counterparts~\citep{liang2025mmactlearnmultimodalparallel,liang2025discretediffusionvlabringing,chen2025unified,wen2025dvla,ye2025dream,liu2026mmada}, Reinforcement Learning (RL) for native dVLAs remains largely underexplored. The primary bottleneck lies in the inherent mismatch between the multi-step denoising process and the exact action likelihood required by standard policy-gradient algorithms. Although a dVLA policy outputs a tractable categorical distribution via standard softmax at each individual denoising step, evaluating the exact marginal probability $\pi_\theta(a_t \mid s_t)$ of the final executed action requires marginalizing over all possible intermediate paths within the iterative, $K$-step Markovian denoising process. This operation is combinatorially expensive; moreover, simple last-step approximations fail to account for the intermediate states that actually dictate the rollout, thereby preventing the direct application of standard RL objectives to dVLA policies.

In this work, we address this gap by formulating RL directly over the denoising path executed by the dVLA. Instead of optimizing the intractable marginal probability of the final action, we unroll the denoising process and express the probability of the sampled generation path as a product of step-wise denoising transitions. This trajectory-level formulation aligns the PPO objective with the actual action-generation process used during rollout and deployment. In practice, each transition consists of a mask-scheduling decision and token-generation probabilities. Since the scheduler is a discrete selection process, we treat it as part of the generation dynamics and compute policy-gradient updates only on the newly denoised action tokens along the executed path. This yields a tractable and stable objective for applying standard PPO to multi-step discrete diffusion VLA policies.

Importantly, this trajectory-level formulation also unifies different denoising horizons. A 1-step decoder can be viewed as a degenerate denoising path with a single transition, while 2-step and 4-step decoding correspond to longer generation paths under the same probability framework. This flexibility enables a hybrid denoising-step strategy: different tasks can use different denoising horizons according to their initial SFT performance and refinement requirements. For tasks where low-step decoding already provides a strong initial success rate, fewer denoising steps reduce rollout and inference cost. For tasks that require more iterative refinement, larger denoising horizons provide better initial actions and more informative online exploration. In this way, the same trajectory-level RL framework can balance action quality, sample efficiency, and computational cost.

Built on these ideas, we present \udvlarl, a reinforcement learning framework for native discrete diffusion VLA policies. \udvlarl enables PPO-style online RL by optimizing the probability of sampled denoising paths rather than approximating the marginal likelihood of final actions. Empirically, we evaluate \udvlarl on LIBERO and RoboTwin 2.0. The results show that \udvlarl consistently improves its SFT-only discrete diffusion VLA backbone, achieves state-of-the-art performance on LIBERO, and obtains strong VLA-based performance on RoboTwin 2.0 while remaining competitive with strong World-Action Model baselines. Further analyses demonstrate the necessity of trajectory-level probability modeling and show that the hybrid denoising-step strategy improves efficiency while preserving the benefits of multi-step action refinement.

In summary, our core contributions are threefold:
\begin{itemize}
\item We formulate reinforcement learning for native discrete diffusion VLAs by optimizing the probability of sampled denoising trajectories instead of the intractable marginal likelihood of final actions. This provides a simple and tractable way to integrate standard policy-gradient RL with multi-step discrete denoising policies.

\item We introduce \udvlarl, a unified denoising-step training strategy enabled by our trajectory-level probability formulation. By assigning task-wise denoising horizons according to task complexity and initial policy performance, our method reduces computational cost on easier tasks while preserving iterative action refinement for harder tasks, improving both effectiveness and efficiency.

\item We empirically demonstrate that reinforcement learning can substantially improve discrete diffusion VLA policies beyond their SFT initialization. Our framework achieves a success rate of \textbf{99.7\%} on LIBERO and delivers a \textbf{30.6\%} improvement over the MM-ACT SFT baseline on RoboTwin 2.0, highlighting the potential of native discrete diffusion as a promising action-generation paradigm for RL-optimized VLA policies, and the effectiveness of \udvlarl.
\end{itemize}

\section{Related Work}
\label{sec:related_work}

\subsection{Diffusion Language, Vision-Language, and Vision-Language-Action Models}
\label{subsec:related_dllm_dvla}

Discrete diffusion language models have recently emerged as an alternative to autoregressive sequence modeling. Instead of generating tokens in a fixed left-to-right order, they construct discrete sequences through masked denoising, enabling parallel prediction, bidirectional conditioning, and iterative refinement~\citep{DBLP:journals/corr/abs-2107-03006,hoogeboom2021argmax,sahoo2024simple,ye2025dream,nie2026large}. This generation paradigm has also been extended to vision-language modeling, where diffusion-based VLMs unify multimodal understanding and generation by denoising visual and textual tokens under shared multimodal contexts~\citep{yang2026mmada,you2026llada}. Compared with autoregressive VLMs, these models provide a more flexible mechanism for order-free multimodal generation and naturally support parallel refinement over multiple token types. Recent discrete diffusion VLA policies further extend this paradigm from language and vision-language generation to robotic action generation~\citep{liang2025mmactlearnmultimodalparallel,liang2025discretediffusionvlabringing,chen2025unified,wen2025dvla,ye2025dream,liu2026mmada}, achieving performance compatible with state-of-the-art VLAs with continuous action modeling~\citep{openvla-oft,intelligence2025pi05visionlanguageactionmodelopenworld}.

However, existing discrete diffusion VLA studies are still largely centered on supervised training and fine-tuning with demonstration data. How to further improve such policies through reinforcement learning remains underexplored. Our method targets this missing piece by formulating discrete diffusion action generation as a trajectory-level policy and enabling PPO-style reinforcement learning over the actually executed denoising process.

\subsection{Reinforcement Learning for Discrete Diffusion Policies}
\label{subsec:related_dllm_rl}

Unlike autoregressive policies, discrete diffusion models do not admit a tractable token-wise decomposition of the marginal likelihood of a generated sequence, since the same final output may be reached through many intermediate denoising paths. Existing reinforcement learning methods largely circumvent this difficulty using endpoint-conditioned denoising surrogates, constructing corrupted versions of completed samples and approximating their policy likelihoods using denoising reconstruction objectives at one or more masking levels~\citep{zhao2025d1scalingreasoningdiffusion,yang2026mmada,gong2026diffucoder}.  While computationally convenient, these objectives evaluate independently constructed denoising states rather than the transition probabilities along the actual inference trajectory, causing cumulative gradient imbalance~\citep{Zhang_2026_CVPR}. This imbalance can be alleviated by reweighting timestep- and token-level gradient contributions according to their masking stages and frequencies, thereby improving optimization stability and training efficiency~\citep{Zhang_2026_CVPR}. Alternatively, intermediate decoding states can be retained and equipped with step-wise value estimation to provide finer-grained credit assignment throughout the denoising process~\citep{wang2025revolutionizingreinforcementlearningframework}.

Discrete diffusion policies for embodied control differ from language and image DDMs in two important aspects. First, dVLAs typically require fewer denoising steps for action generation, making explicit modeling of the sampled denoising path practical~\citep{liang2025mmactlearnmultimodalparallel}. Second, intermediate action-token sequences are merely transient refinement states with limited standalone semantics; only the fully denoised action is executed and interacts with the environment. Consequently, treating each denoising step as an independent environment-level decision introduces unnecessary credit estimation for intermediate action proposals that serve only as refinement states and receive no direct environmental feedback. Our method instead treats the short denoising chain as an internal action-generation trajectory nested within a single environment decision. The joint likelihood of the sampled trajectory factorizes tractably over its Markov transitions and is directly used for policy optimization, avoiding both endpoint-conditioned surrogates and auxiliary step-wise value estimation while remaining consistent with the generation process used during rollout and deployment.

\subsection{Reinforcement Learning for Vision-Language-Action Policies}
\label{subsec:related_vla_rl}

Vision-Language-Action policies are traditionally developed through Supervised Fine-Tuning on expert demonstrations. However, pure behavior cloning is susceptible to compounding errors during closed-loop interaction, fundamentally capping the upper bound of task success. To overcome this bottleneck, an emerging body of work has introduced Reinforcement Learning to VLA training, shifting the optimization objective from mimicking static datasets to maximizing environmental rewards. This paradigm shift has primarily evolved across two distinct action generation paradigms. On the one hand, RL has been extensively integrated into mainstream VLAs with discrete action tokenization, where researchers have successfully adapted LLM-style policy optimization algorithms, and built efficient system infrastructures~\citep{lu2025vlarlmasterfulgeneralrobotic, tan2025interactiveposttrainingvisionlanguageactionmodels, chen2025tgrpofinetuningvisionlanguageactionmodel, yu2025rlinf, zang2026rlinfvlaunifiedefficientframework, li2026simplevlarl}. On the other hand, for continuous-action policies, offline methods such as Flow Q-Learning
avoid directly optimizing through iterative flow generation by learning
a one-step actor~\citep{park2025flowqlearning}. Online methods for
flow-based generation introduce tractable stochastic formulations,
including ODE-to-SDE conversion or auxiliary noise-transition models
~\citep{liu2025flowgrpotrainingflowmatching,
chen2026pitextttrlonlinerlfinetuning}. Other approaches, such as RL Token,
attach a lightweight actor--critic interface to a pretrained VLA rather
than directly optimizing the likelihood of its original action generator
~\citep{xu2026rltokenbootstrappingonline}.

Despite this rapid proliferation across mainstream discrete and continuous paradigms, the application of RL to native discrete diffusion VLA policies remains largely unexplored. Our method makes the first attempt to apply reinforcement learning to this specific paradigm.
\section{Preliminaries}
\label{sec:preliminary}

In this section, we establish the mathematical formalisms for the two foundational components of our framework: the discrete diffusion model for action generation and the standard Markov Decision Process for robotic control.

\subsection{Discrete Diffusion Models}
\label{subsec:discrete_diffusion}

In the context of Vision-Language-Action models, we define $x_k$ strictly as the discrete action sequence at denoising step $k \in \{K, K-1, \dots, 0\}$, where a subset of the tokens is replaced by a special \texttt{[MASK]} token. The generation process is conditioned on the state $s$, which comprises visual observations, language instructions, and robot proprioception.

The forward process progressively corrupts the ground-truth sequence $x_0$ by replacing tokens with \texttt{[MASK]} according to a predefined schedule. Conversely, the generative process is a parameterized Markov chain that reconstructs the sequence from a fully masked state $x_K$ to the fully decoded state $x_0$. At each denoising step $k$, a neural network parameterized by $\theta$ predicts the categorical probability distribution of the masked tokens, conditioned on the current sequence and the state $s$, denoted as $p_\theta(\cdot | x_k, s)$. Subsequently, a scheduler then determines the specific subset of tokens $U_k$ to be decoded or re-masked based on the network's confidence.

\subsection{Standard MDP and PPO Objective}
\label{subsec:mdpandppo}

We formulate the robotic control task as a discrete-time Markov Decision Process defined by the tuple $\langle \mathcal{S}, \mathcal{A}, \mathcal{P}, \mathcal{R}, \gamma \rangle$. At each environmental timestep $t$, the agent receives a state $s_t \in \mathcal{S}$ and executes an action $a_t \in \mathcal{A}$. In our VLA framework, the final action $a_t$ exactly corresponds to the fully decoded sequence $x_0$ generated by the discrete diffusion model. The environment then yields a reward $r_t = \mathcal{R}(s_t, a_t)$ and transitions to the next state.

The agent's behavior is governed by a policy $\pi_\theta(a_t | s_t)$. To optimize this policy, Proximal Policy Optimization maximizes the clipped surrogate objective:
\begin{equation}
    L^{\mathrm{CLIP}}(\theta) = \mathbb{E}_{t} \left[ \min \left( r_t(\theta) A_t, \mathrm{clip}(r_t(\theta), 1-\epsilon, 1+\epsilon) A_t \right) \right]
\end{equation}
where $A_t$ is the advantage estimate, and $r_t(\theta)$ denotes the importance sampling ratio:
\begin{equation}
    r_t(\theta) = \frac{\pi_\theta(a_t | s_t)}{\pi_{\theta_{\mathrm{old}}}(a_t | s_t)}
\end{equation}

\textbf{The intractability dilemma:} Policy gradient algorithms like PPO inherently require evaluating the exact action probability $\pi_\theta(a_t | s_t)$. While this evaluation is computationally trivial for conventional policies that generate actions via a single forward pass, it becomes fundamentally intractable for discrete diffusion models. Because the final action $a_t$ emerges from a $K$-step stochastic denoising process, calculating its exact marginal probability requires marginalizing over the combinatorial space of all possible intermediate denoising trajectories $\tau = (x_K, x_{K-1}, \dots, x_0)$ that culminate in $a_t$:
\begin{equation}
    \pi_\theta(a_t | s_t) = \sum_{x_1, \dots, x_K} p(x_K) \prod_{k=1}^K p_\theta(x_{k-1} | x_k, s_t).
\end{equation}
This exact marginalization is computationally intractable due to the combinatorial explosion of valid transition paths across the $K$-step discrete denoising process. 

Furthermore, treating the network's prediction at any single intermediate step (e.g., $p_\theta(\cdot | x_k, s_t)$) as a proxy for the overall policy probability is fundamentally flawed. Such a single-step output merely provides a local conditional probability given a transient noisy state $x_k$. It fails to account for the joint mathematical product of the entire Markov chain, where each intermediate transition is iteratively governed by the scheduling mechanism. In Section~\ref{sec:method}, we resolve this dilemma by unrolling the sequential process and formulating the PPO objective directly over the joint probability of the sampled trajectory, thereby gracefully bypassing the intractable marginalization.
\begin{figure}[htbp]
    \centering
    \includegraphics[width=\linewidth]{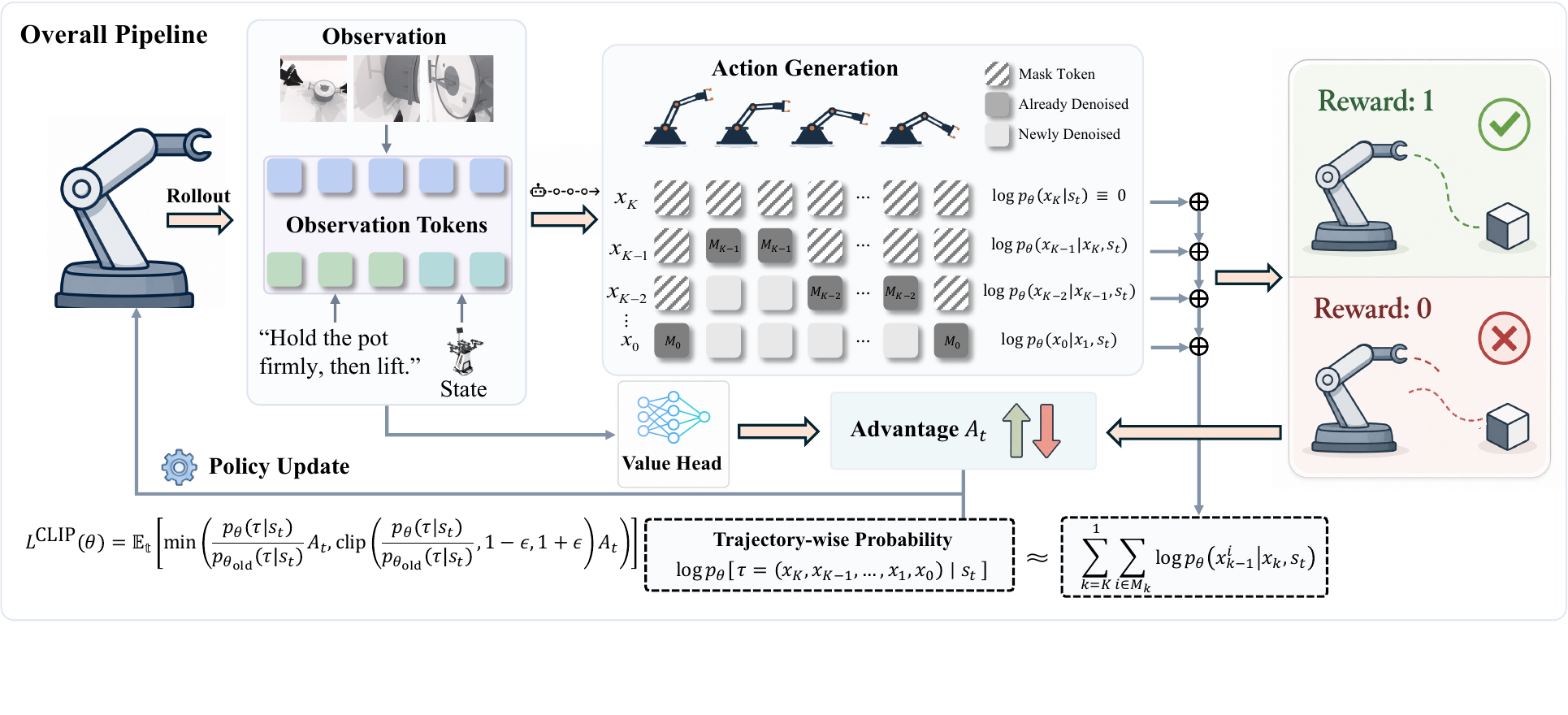}
    \caption{\textbf{The overall pipeline of \udvlarl.} (1) \textbf{Rollout \& Action Generation:} During interaction, the agent conditions on multimodal observations $s_t$ to generate actions via a $K$-step iterative masked denoising process, executing them to collect trajectory rewards. (2) \textbf{Trajectory-Level Policy Optimization:} To bypass intractable marginals, we formulate the joint probability of the unrolled denoising path, $\log p_{\theta}(\tau|s_t)$, by summing the log-likelihoods of actively unmasked tokens. Combined with advantages $A_t$ estimated by the value head, this trajectory-wise likelihood enables stable and efficient updates via the standard PPO clipped surrogate objective.}
    \label{fig:pipeline}
\end{figure}

\section{Method}
\label{sec:method}

\subsection{Modeling the Joint Probability of Markovian Denoising Trajectories}
\label{subsec:joint_prob}

To resolve the intractability dilemma outlined in Section~\ref{subsec:mdpandppo}, we propose shifting the optimization objective from the intractable marginal action probability $\pi_\theta(a_t | s_t)$ to the joint probability of a specifically sampled denoising trajectory $\pi_\theta(\tau | s_t)$. Unlike autoregressive models that deterministically factorize the marginal probability of a sequence, discrete diffusion generation intrinsically operates as a parameterized Markov chain over $K$ steps. By unrolling this sequential process and leveraging the Markov property, the joint probability of a complete trajectory $\tau = (x_K, \dots, x_0)$ smoothly factorizes into the product of single-step transitions.

Specifically, at each step $k$, the transition from $x_k$ to $x_{k-1}$ is decoupled into two distinct mechanisms: token unmasking and token generation. The single-step transition probability is formulated as:
\begin{equation}
    P_{\theta}(x_{k-1}|x_k, s_t) = P_{\mathrm{unmask}}(U_k | p_\theta, k) \prod_{i \in U_k} p_\theta(x_{k-1}^i | x_k, s_t),
\end{equation}
where $U_k$ denotes the specific subset of token indices selected by the scheduler to be updated at step $k$, and $P_{\mathrm{unmask}}(U_k | p_\theta, k)$ is the probability of this selection, driven by the network's output confidence $p_\theta$ via a Gumbel-TopK mechanism. Since the transition for unselected tokens ($j \notin U_k$) is deterministic with $P(x_{k-1}^j | x_k^j) \equiv 1$, this term is naturally omitted from the product.

Consequently, by aggregating these single-step transitions across the entire unrolled process, we establish our tractable trajectory-level policy. The log-probability of generating the specific path $\tau$ is thus strictly formulated as:
\begin{equation}
    \log \pi_\theta(\tau | s_t) = \sum_{k=K}^1 \left[ \log P_{\mathrm{unmask}}(U_k | p_\theta, k) + \sum_{i \in U_k} \log p_\theta(x_{k-1}^i | x_k, s_t) \right].
\end{equation}

\subsection{Implicit optimization of the unmasking process}
\label{subsec:implicit optimization}

By redefining the PPO ratio over the trajectory-level policy, $\log r_t(\theta) = \log \pi_{\theta}(\tau|s_t) - \log \pi_{\theta_{\mathrm{old}}}(\tau|s_t)$, the objective naturally decomposes into a token unmasking difference term and a token generation difference term. However, retaining the token unmasking term for gradient updates introduces severe optimization pathologies.

First, under PPO's trust region constraint ($\theta \approx \theta_{\mathrm{old}}$), the selection probabilities of the currently decoded tokens remain largely invariant, rendering their gradient contributions numerically negligible. Second, and more critically, the Gumbel-TopK scheduler relies on discrete sorting operations. Attempting to backpropagate through this term introduces non-differentiable step functions, generating high-variance gradient impulses that can easily violate the PPO clip constraint and severely destabilize training.

To resolve this, we treat the unmasking process as an intrinsic, non-differentiable system dynamic. By dropping the token unmasking term, our tractable surrogate objective focuses exclusively on token generation. Furthermore, to eliminate gradient noise from already-decoded tokens, we introduce a masked objective. We restrict the gradient computation strictly to the subset of tokens $M_k \subseteq U_k$ that undergo a phase transition from the \texttt{[MASK]} state to a specific token ID:
\begin{equation}
    L^{\mathrm{\udvlarl}}(\theta) = \mathbb{E}_{\tau} \left[ \sum_{k=K}^1 \sum_{i \in M_k} \log p_\theta(x_{k-1}^i | x_k, s_t) \cdot A_t \right].
\end{equation}
where $A_t$ is a chunk-level advantage, which uniformly weights the unrolled generation steps.

Although the explicit loss is exclusively computed on the subset $M_k$, the network confidence $p_\theta$ is shared across both decoupled mechanisms. As the policy gradient improves the core token prediction accuracy, the global contextual representation is enhanced, causing the overall distribution of $p_\theta$ to become increasingly confident and calibrated. This implicitly and adaptively optimizes the forward pass of the Gumbel-TopK scheduler, allowing it to accurately target highly uncertain regions without requiring explicit, high-variance gradient estimators.
\section{Experiments}
\label{sec:experiments}

We organize our experiments around four research questions. First, we examine whether \udvlarl can improve dVLA policies, and how the resulting policy compares with both its SFT-only backbone and existing VLA and WAM baselines. Second, we investigate whether trajectory-level probability modeling is necessary for RL on dVLA policies by comparing it with a final-step-only proxy under the same rollout and reward settings. Third, we analyze how different denoising-trajectory lengths affect action quality, initial success rates, RL optimization, and sample efficiency. Building on the unified trajectory-level formulation, we further introduce Hybrid \udvlarl, which assigns different denoising steps to different tasks within the same RL framework. Finally, we evaluate whether Hybrid \udvlarl improves training and inference efficiency while preserving the benefits of multi-step action refinement and effective online exploration.

\subsection{Experimental Setup}
\label{subsec:exp_setup}

\paragraph{Benchmarks.}
We evaluate \udvlarl on two widely used robotic manipulation simulation benchmarks: LIBERO~\citep{liu2023liberobenchmarkingknowledgetransfer} and RoboTwin 2.0~\citep{chen2025robotwin20scalabledata}. 
LIBERO consists of four standard single-arm manipulation suites, Spatial, Object, Goal, and Long, which evaluate spatial relations, object-centric manipulation, goal-conditioned variation, and long-horizon compositional instructions. 
RoboTwin 2.0 is a more challenging bimanual manipulation benchmark with diverse objects, domain-randomized environments, and dual-arm coordination.

\paragraph{Backbone and SFT initialization.}
We instantiate \udvlarl on MM-ACT~\citep{liang2025mmactlearnmultimodalparallel}, a unified discrete diffusion VLA model that represents action chunks as discrete tokens and generates them through masked denoising. 
For fair comparison with existing VLA-RL studies, we use the same scale of SFT demonstrations as SimpleVLA-RL~\citep{li2026simplevlarl}: 500 demonstrations per suite on LIBERO and 1,000 demonstrations per task on RoboTwin 2.0. 
Different from SimpleVLA-RL, which fine-tunes policies for individual tasks or suites, our LIBERO and RoboTwin 2.0 SFT checkpoints are obtained through multi-task joint training. 
Thus, the SFT initialization used by \udvlarl is a unified multi-task dVLA policy rather than a task-specific policy. 

\paragraph{Online RL and evaluation protocol.}
Starting from the SFT checkpoint, we conduct online RL post-training in simulation using RLinf~\citep{yu2025rlinf}. 
We adopt the PPO~\citep{SchulmanWDRK17} implementation in RLinf, while using \udvlarl to compute trajectory-level denoising probabilities for policy optimization. 
The reward is a sparse outcome reward: a trajectory receives reward $1$ if the task is completed successfully and $0$ otherwise. 
Unless otherwise specified, we report success rate (SR) as the primary metric. For SFT checkpoint evaluation, we evaluate each LIBERO subtask with 50 trials, resulting in 2,000 trials over the four LIBERO suites. 
For RoboTwin 2.0, the SFT checkpoint is evaluated under 4-step denoising with 100 trials per task. 
For \udvlarl, the success rates reported in the main tables correspond to the best online rollout success rates observed during RL training. 
On LIBERO, we report the best task-wise online SR observed during training, where each task is evaluated with 512 rollout episodes. 
On RoboTwin 2.0, each task uses 1,000 training scenarios for online RL; we report the highest average SR across the eight selected tasks and the corresponding per-task SRs, using 64 rollout episodes per task. More details are provided in Appendix~\ref{app:libero_rl_training_details} and Appendix~\ref{app:robotwin_rl_training_details}.

\paragraph{Baselines and comparisons.}
We compare \udvlarl with representative VLA and World Action Model (WAM) baselines, as well as with its SFT-only backbone, denoted as \textsc{MM-ACT}$^\ast$. 
\textsc{MM-ACT}$^\ast$ adopts the same architecture and SFT initialization as \udvlarl but does not perform online RL post-training; therefore, this comparison isolates the effect of reinforcement learning. 
Compared with the original MM-ACT~\citep{liang2025mmactlearnmultimodalparallel}, \textsc{MM-ACT}$^\ast$ uses action-only SFT and an action chunk size of 16 for improved training and inference efficiency. 
In addition, while the original MM-ACT trains separate models for the four LIBERO suites, \textsc{MM-ACT}$^\ast$ is jointly trained across all suites. 
Other implementation details are provided in Appendix~\ref{app:sft_details}. 
Comparisons with other VLA methods and WAMs further contextualize whether RL-optimized dVLAs achieve competitive performance against existing manipulation policies.

\subsection{Does \udvlarl Improve Discrete Diffusion VLA Policies?}
\label{subsec:main_results}

We examine this question by comparing \udvlarl with its SFT-only backbone and existing baselines on LIBERO and RoboTwin 2.0, covering both single-arm and bimanual manipulation.

\paragraph{LIBERO main results.}
Table~\ref{tab:libero_main} reports the main results on LIBERO. 
\udvlarl achieves state-of-the-art performance on LIBERO, obtaining success rates above 99\% across all four suites, reaching an average success rate of \textbf{99.7\%}. 
Compared with the SFT-only MM-ACT backbone, \udvlarl brings clear improvements, showing that online RL provides effective task-level supervision beyond behavior cloning. 
Moreover, \udvlarl outperforms representative VLA- and WAM-based baselines in average success rate, demonstrating that discrete diffusion VLA policies can reach state-of-the-art manipulation performance when their denoising-based action generation process is optimized with the proposed trajectory-level probability formulation.

\begin{table}[t]
\caption{Main results on LIBERO simulation benchmark. We report per suite success rate (\%) and average success rate across 4 suites.}
\label{tab:libero_main}
\centering
\small
\setlength{\tabcolsep}{4pt}
\renewcommand{\arraystretch}{1.08}
\begin{tabular}{lccccc}
\hline
Method & Spatial & Object & Goal & Long & Avg. \\
\hline

\multicolumn{6}{l}{\textbf{\# VLA-Based}} \\
OpenVLA~\citep{openvla}
& 84.7 & 88.4 & 79.2 & 53.7 & 76.5 \\
$\pi_0$~\citep{pi_0}
& 96.8 & 98.8 & 95.8 & 85.2 & 94.2 \\
UniVLA~\citep{bu2025univlalearningacttaskcentric}
& 96.5 & 96.8 & 95.6 & 92.0 & 95.2 \\
$\pi_{0.5}$~\citep{intelligence2025pi05visionlanguageactionmodelopenworld}
& 98.8 & 98.2 & 98.0 & 92.4 & 96.9 \\
OpenVLA-OFT~\citep{openvla-oft}
& 97.6 & 98.4 & 97.9 & 94.5 & 97.1 \\
SimpleVLA-RL~\citep{li2026simplevlarl}
& 99.4 & 99.1 & 99.2 & 98.5 & 99.1 \\
MM-ACT~\citep{liang2025mmactlearnmultimodalparallel}
& 97.8 & 99.4 & 94.8 & 93.0 & 96.3 \\
MM-ACT*
& 91.2 & 82.6 & 90.0 & 88.7 & 88.1 \\
\udvlarl \textbf{(Ours)}
& \textbf{99.8} & \textbf{100.0} & \textbf{99.6} & \textbf{99.2} & \textbf{99.7} \\

\hline
\multicolumn{6}{l}{\textbf{\# WAM-Based}} \\
Motus~\citep{bi2026motus}
& 96.8 & 99.8 & 96.6 & 97.6 & 97.7 \\
Cosmos Policy~\citep{kim2026cosmos}
& 98.1 & 100.0 & 98.2 & 97.6 & 98.5 \\
LingBot-VA~\citep{li2026causalworldmodelingrobot}
& 98.5 & 99.6 & 97.2 & 98.5 & 98.5 \\
Fast-WAM~\citep{yuan2026fastwamworldactionmodels}
& 98.2 & 100.0 & 97.0 & 95.2 & 97.6 \\
\hline
\end{tabular}
\end{table}

\paragraph{RoboTwin 2.0 main results.}
Table~\ref{tab:robotwin_per_task} reports the per-task success rates on the selected RoboTwin 2.0 tasks. 
The results show that the improvements of \udvlarl are not limited to single-arm manipulation. 
Compared with the SFT-only backbone, \udvlarl consistently improves performance across tasks involving grasping, placing, object transfer, and tool-object interaction, suggesting that the proposed trajectory-level RL formulation also benefits more complex bimanual settings that require coordinated control and longer closed-loop execution.

Notably, \udvlarl achieves state-of-the-art performance among VLA-based methods on RoboTwin 2.0.
Compared with the SFT-only \textsc{MM-ACT}$^\ast$ backbone, \udvlarl improves the average success rate from $61.4\%$ to $92.0\%$, corresponding to an absolute gain of $30.6$ percentage points.
The improvement is observed across all eight tasks, with particularly large gains on Handover Mic ($+47.9$ points), Move Can Pot ($+47.5$ points), and Lift Pot ($+43.1$ points).
These results indicate that trajectory-level RL substantially strengthens the SFT policy on challenging bimanual tasks involving coordinated control and long-horizon execution.

RoboTwin 2.0 is also a visually challenging benchmark with strong domain randomization.
For completeness, we report representative WAM-based results as reference points~\citep{bi2026motus,ye2026gigaworldpolicyefficientactioncenteredworldaction,li2026causalworldmodelingrobot,yuan2026fastwamworldactionmodels}.
However, these methods follow different RoboTwin training settings and therefore should not be interpreted as strictly controlled comparisons with \udvlarl.
Under this caveat, the WAM-based results provide additional context for assessing the absolute performance level of \udvlarl on RoboTwin 2.0.
The strong improvement over its SFT-only backbone and its leading performance among VLA-based methods further demonstrate the potential of native dVLA policies and the effectiveness of trajectory-level RL optimization.

Overall, the results on LIBERO and RoboTwin 2.0 show that \udvlarl consistently improves the SFT-only dVLA policy. 
It achieves state-of-the-art performance on LIBERO, delivers favorable results compared with other VLA-based methods on RoboTwin 2.0, and remains competitive with strong WAM-based policies. This directly supports the effectiveness of applying trajectory-level RL optimization to discrete diffusion action generation.

\begin{table*}[t]
\caption{Per-task results on RoboTwin 2.0. We report success rate (\%) on the eight selected tasks. Gray rows denote WAM-based reference results.}
\label{tab:robotwin_per_task}
\centering
\scriptsize
\setlength{\tabcolsep}{3.2pt}
\renewcommand{\arraystretch}{1.08}
\newcommand{\taskcell}[1]{%
    \parbox[c][3.6em][c]{0.0865\textwidth}{\centering #1}%
}
\begin{tabular}{lccccccccc}
\hline
Method
& \taskcell{Beat Block Hammer}
& \taskcell{Place Phone Stand}
& \taskcell{Pick Dual Bottles}
& \taskcell{Lift Pot}
& \taskcell{Move Can Pot}
& \taskcell{Place A2B Left}
& \taskcell{Place Empty Cup}
& \taskcell{Handover Mic}
& Avg. \\
\hline

\multicolumn{10}{l}{\textbf{\# VLA-Based}} \\
$\pi_0$
& 59.0 & 22.0 & 50.0 & 51.0 & 41.0 & 38.0 & 60.0 & 96.0 & 52.1 \\
RDT
& 22.0 & 13.0 & 18.0 & 45.0 & 33.0 & 21.0 & 42.0 & 95.0 & 36.1 \\
OpenVLA-OFT
& 28.1 & 17.1 & 29.7 & 10.1 & 28.1 & 37.5 & 77.3 & 45.3 & 34.2 \\
SimpleVLA-RL
& 87.5 & 39.6 & 68.3 & 64.1 & 61.2 & 45.3 & 94.2 & 89.2 & 68.7 \\
$\pi_{0.5}$
& 88.0 & 73.0 & 98.0 & 88.0 & 72.0 & 74.0 & 74.0 & 32.0 & 74.9 \\
MM-ACT*
& 85.0 & 62.0 & 71.0 & 46.0 & 40.0 & 57.0 & 81.0 & 49.0 & 61.4 \\
\udvlarl \textbf{(Ours)}
& \textbf{95.3} & \textbf{96.9} & \textbf{95.3} & \textbf{89.1}
& \textbf{87.5} & \textbf{79.7} & \textbf{95.3} & \textbf{96.9}
& \textbf{92.0} \\

\hline
\multicolumn{10}{l}{\textbf{\# WAM-Based}} \\
\rowcolor{gray!15}
Motus
& 88.0 & 86.0 & 90.0 & 99.0 & 74.0 & 79.0 & 98.0 & 63.0 & 84.6 \\
\rowcolor{gray!15}
GigaWorld
& 86.0 & 72.0 & 86.0 & 98.0 & 78.0 & 88.0 & 90.0 & 72.0 & 83.8 \\
\rowcolor{gray!15}
LingBot-VA
& 98.0 & 97.0 & 99.0 & 99.0 & 97.0 & 93.0 & 100.0 & 96.0 & 97.4 \\
\rowcolor{gray!15}
Fast-WAM
& 97.0 & 99.0 & 96.0 & 100.0 & 88.0 & 93.0 & 100.0 & 100.0 & 96.6 \\
\hline
\end{tabular}
\end{table*}

\subsection{Is Trajectory-Level Probability Modeling Necessary for Discrete Diffusion VLA RL?} 
\label{subsec:trajectory_modeling} 
This subsection examines whether trajectory-level probability modeling is necessary for RL on dVLA policies. Unlike conventional VLA policies that output an action chunk through a single forward prediction, a dVLA generates actions through an iterative denoising process. At each denoising step, the policy predicts token distributions conditioned on the current partially denoised action sequence, and the final executable action is produced only after the complete denoising trajectory has been executed. Therefore, the policy likelihood used for RL should faithfully reflect the probability of the action-generation path followed during rollout, rather than treating the final action as an isolated prediction. \udvlarl formulates this process as a Markovian denoising trajectory and optimizes the probability of the executed greedy denoising path, aligning the RL objective with the actual action-generation process used at deployment.

\begin{figure}[t]
\centering
\begin{subfigure}{0.49\linewidth}
\centering
\includegraphics[width=\linewidth]{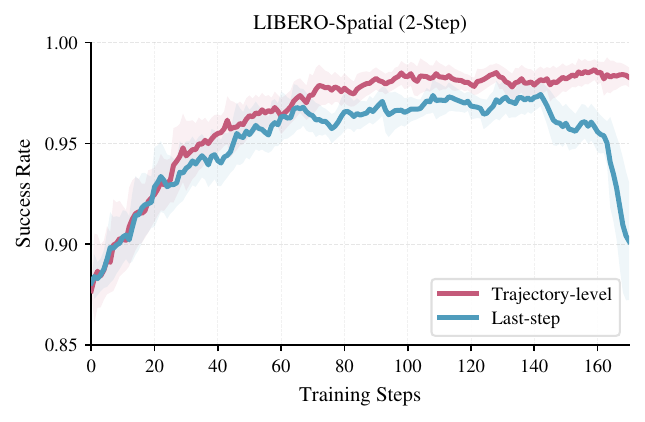}
\label{fig:libero_spatial_2step_logp}
\end{subfigure}
\hfill
\begin{subfigure}{0.49\linewidth}
\centering
\includegraphics[width=\linewidth]{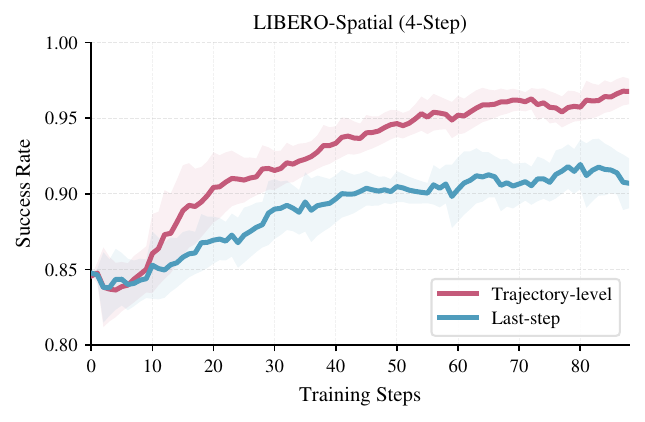}
\label{fig:libero_spatial_4step_logp}
\end{subfigure}

\vspace{0.5em}

\begin{subfigure}{0.49\linewidth}
\centering
\includegraphics[width=\linewidth]{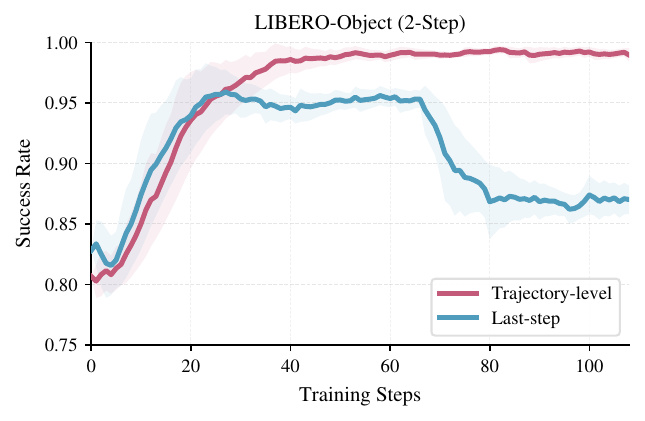}
\label{fig:libero_object_2step_logp}
\end{subfigure}
\hfill
\begin{subfigure}{0.49\linewidth}
\centering
\includegraphics[width=\linewidth]{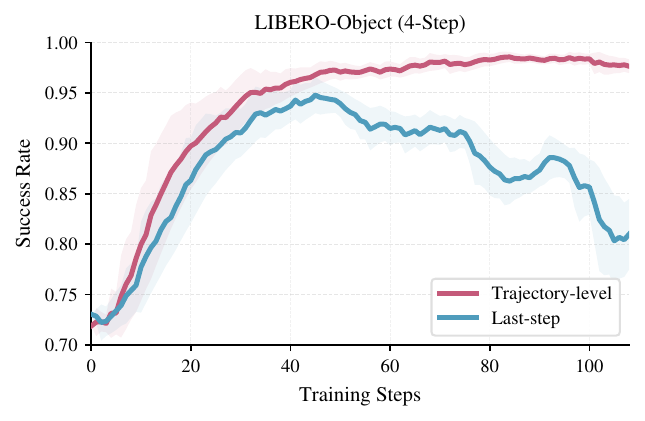}
\label{fig:libero_object_4step_logp}
\end{subfigure}

\caption{
Comparison between the trajectory-level log-probability objective and the last-step log-probability approximation on LIBERO-Spatial and LIBERO-Object under different denoising steps. Solid lines represent the Exponential Moving Average (EMA) of the success rate, and the shaded regions indicate the corresponding $\pm 1$ standard deviation.}
\label{fig:libero_logp_comparison}
\end{figure}

\paragraph{Necessity of trajectory-level optimization.} On LIBERO-Spatial and LIBERO-Object, we compare the proposed trajectory-level objective with a final-step-only proxy under both 2-step and 4-step denoising. The final-step-only variant approximates the policy likelihood using only the log-probability from the last denoising step, while ignoring the intermediate denoising transitions that lead to the final action. Since the two variants share the same backbone, rollout setting, and sparse outcome reward, their difference isolates the effect of trajectory-level probability modeling. As shown in Figure~\ref{fig:libero_logp_comparison}, the trajectory-level objective consistently yields more stable improvement across tasks and denoising horizons. On LIBERO-Spatial, the final-step-only proxy can improve in the early stage but becomes less stable later in training, especially under 2-step denoising, whereas the trajectory-level objective maintains a higher and more stable success rate. Under 4-step denoising, the trajectory-level objective continues to improve steadily, while the final-step-only proxy plateaus at a lower success rate. A similar trend is observed on LIBERO-Object. Although both variants initially benefit from RL, the final-step-only proxy later degrades markedly, while the trajectory-level objective remains near saturation under both 2-step and 4-step denoising. These results indicate that intermediate denoising transitions are not merely implementation details. Ignoring them leads to a less faithful policy likelihood for the executed denoising path, and this mismatch becomes more harmful when action generation involves additional intermediate states. Therefore, modeling the full denoising trajectory is important for stable policy optimization in dVLA policies.

\subsection{How Do Denoising Steps Affect RL Optimization?} \label{subsec:denoising_steps_hybrid} 
The previous subsection shows that the denoising trajectory should be modeled as part of the policy likelihood. We now analyze how the length of this trajectory affects action generation and RL optimization. Importantly, the trajectory-level formulation does not restrict \udvlarl to a fixed denoising schedule. Since different denoising-step settings correspond to trajectories of different lengths, they can be optimized within the same probability framework by evaluating the corresponding executed denoising path. This unified view naturally enables a hybrid denoising-step strategy. Therefore, we further introduce Hybrid \udvlarl, which assigns denoising steps according to task difficulty and the initial capability of the SFT policy. For tasks where low-step decoding already provides a strong initial success rate, using fewer denoising steps can reduce rollout and inference cost without substantially sacrificing exploration quality. For tasks that require more refinement, larger denoising steps can provide better initial actions and expose RL to more successful trajectories under sparse rewards. Detailed task-specific denoising-step configurations are provided in Appendix~\ref{app:robotwin_rl_training_details}.In this way, Hybrid \udvlarl balances sample efficiency with training and inference efficiency under a unified trajectory-level probability framework.

\paragraph{Effect of different denoising steps.}
From the trajectory-level perspective, different denoising horizons correspond to different factorizations of the action-generation probability.
With 1-step denoising, action generation collapses into a single transition, and the corresponding policy likelihood reduces to the one-step parallel-decoding objective.
In contrast, 2-step denoising introduces an intermediate state, while 4-step denoising further unrolls the generation process into a longer trajectory with finer-grained token refinement.
These additional denoising transitions can improve the quality of the action chunks generated by the SFT policy, resulting in stronger initial performance before RL post-training.

\begin{figure}[h]
    \begin{center}
    \includegraphics[width=0.78\linewidth]{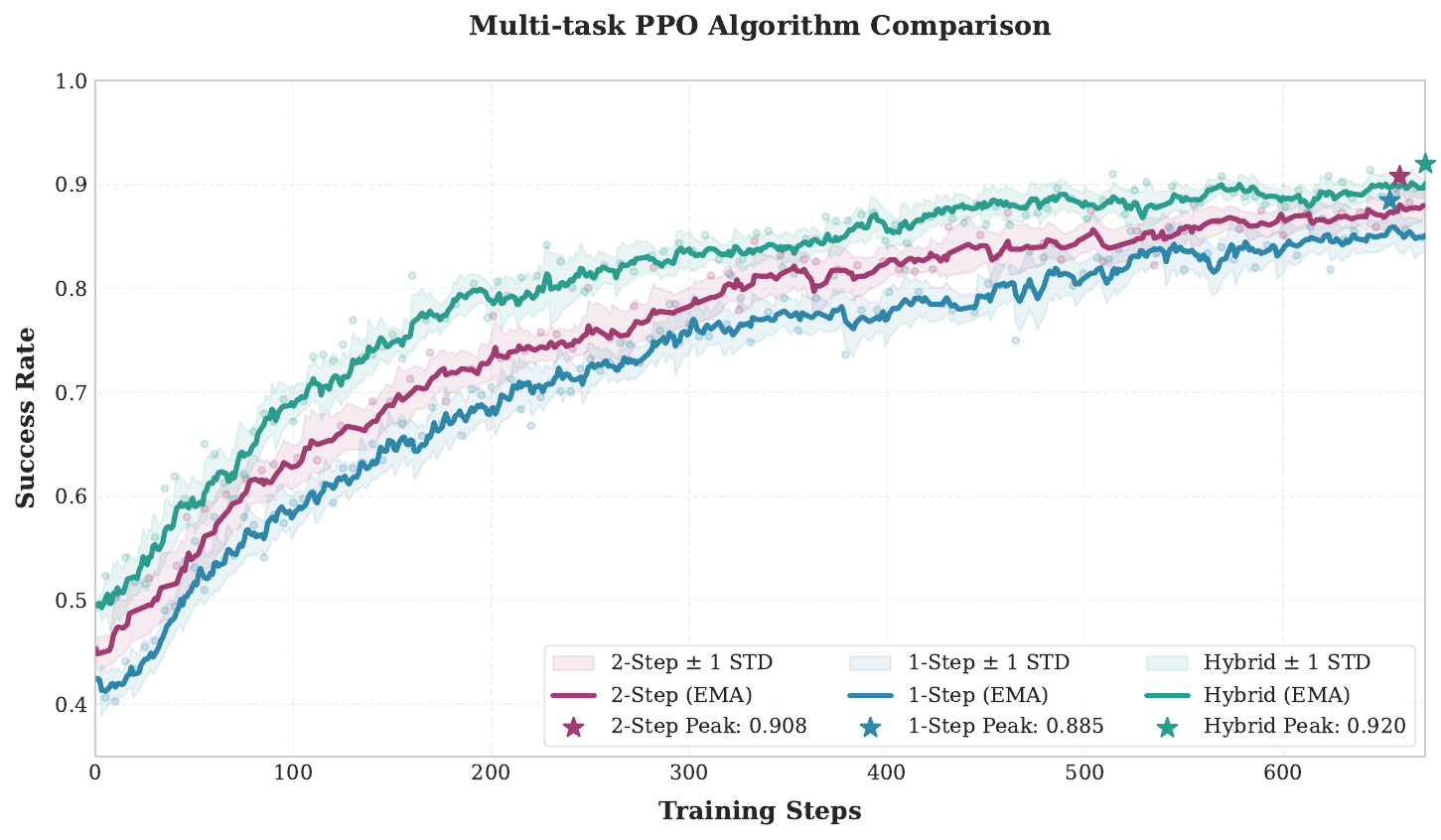}
    \end{center}
    \caption{Effect of denoising-trajectory length on RL optimization in RoboTwin 2.0. Step-1 collapses action generation into a single transition, while Step-2 and Step-4 provide longer denoising trajectories. Hybrid combines Step-1, Step-2, and Step-4 during RL. The curves show that multi-step and mixed-step variants achieve more favorable optimization behavior than fixed one-step decoding.}
    \label{fig:robotwin_hybrid_curve}
\end{figure}

A stronger initialization is particularly beneficial under sparse outcome rewards.
When the initial policy succeeds more frequently, online RL receives informative positive feedback earlier and can improve more rapidly under the same training budget.
As shown in Figure~\ref{fig:robotwin_hybrid_curve}, the fixed multi-step variant starts from a higher success rate and improves faster than fixed 1-step parallel decoding during the early and intermediate stages of training.
Hybrid \udvlarl exhibits the most favorable learning behavior: it maintains the highest success rate over most of the optimization process and reaches the high-success-rate regime earlier than the fixed-step alternatives.
Under the same number of training steps, Hybrid and multi-step denoising therefore achieve higher success rates than 1-step decoding, indicating better sample efficiency.
Hybrid also attains the highest peak success rate of $0.920$, compared with $0.908$ for fixed multi-step denoising and $0.885$ for 1-step decoding.

Rather than using a single denoising horizon for all tasks, Hybrid \udvlarl assigns 1-step, 2-step, or 4-step denoising according to the initial SFT performance of each task.
This task-wise assignment preserves iterative refinement where it is beneficial while avoiding unnecessary denoising transitions on tasks that can already be solved effectively with fewer steps.
Overall, the results show that the denoising horizon affects RL through two coupled mechanisms: it determines the degree of iterative action refinement before and during RL, and it defines the trajectory structure over which policy probabilities are evaluated and optimized.
By adapting this horizon across tasks, Hybrid \udvlarl achieves a more favorable balance among initial policy quality, sample efficiency, and attainable task performance.

\subsection{Does Hybrid Denoising Steps Improve Training and Inference Efficiency?} \label{subsec:hybrid_efficiency} 
Our final analysis examines whether Hybrid \udvlarl can improve computational efficiency while preserving effective online exploration. Fixed multi-step denoising provides stronger action refinement and often leads to a better initial success rate, but it also increases training and inference cost. In contrast, 1-step denoising requires less per-step computation, but may require more online exploration when the initial policy is less successful. Hybrid \udvlarl addresses this trade-off by leveraging the unified trajectory-level probability framework: it uses low-step denoising for tasks where the SFT policy already performs well, and uses more denoising steps for tasks that benefit from additional refinement. Concretely, we evaluate the SFT checkpoint under different denoising steps and select the smallest denoising-step setting that achieves a success rate comparable to longer-step decoding.

\begin{table}[t]
\centering
\small
\setlength{\tabcolsep}{6pt}
\caption{
Efficiency comparison of different denoising-step settings. 
Training time measures the wall-clock cost of RL training, while inference delay and inference FLOPs measure the deployment-time cost of generating one action chunk.
}
\label{tab:hybrid_efficiency}
\begin{tabular}{lccc}
\toprule
Setting & Training time (s) & Inference delay (ms) & Inference FLOPs (GFLOPs) \\
\midrule
1-Step  & 259.33 & 182.47 & 17547.82 \\
2-Step  & 452.06 & 364.21 & 35095.64 \\
4-Step  & 845.95 & 728.92 & 70191.28 \\
Hybrid  & 607.89 & 387.24 & 37289.12 \\
\bottomrule
\end{tabular}
\end{table}

Table~\ref{tab:hybrid_efficiency} reports the training and inference efficiency of different denoising-step settings. As expected, increasing the number of denoising steps improves action refinement but introduces nearly proportional computational overhead during both rollout collection and deployment. Hybrid \udvlarl reduces this burden by avoiding fixed long-step denoising for all tasks. Instead, it mixes 1-step, 2-step, and 4-step settings according to task difficulty, retaining the stronger initial performance of multi-step denoising while keeping the average training and inference cost lower than always using 4-step decoding. These results suggest that Hybrid \udvlarl improves both sample efficiency and computational efficiency, making trajectory-level RL more practical for large-scale online rollout collection and low-latency robotic deployment.
\section{Conclusion}
\label{sec:conclusion}

In this work, we presented \udvlarl, a reinforcement learning framework for discrete diffusion Vision-Language-Action policies. Instead of approximating the intractable marginal likelihood of the final action, \udvlarl formulates action generation as an unrolled Markovian denoising trajectory and optimizes its tractable trajectory-level probability. To make this objective stable for policy-gradient optimization, we treat the discrete scheduler as an intrinsic system dynamic and restricts gradient updates to actively generated tokens. This design enables standard PPO-style reinforcement learning to directly optimize the denoising process executed at inference time.

Empirically, \udvlarl consistently improves its SFT-only discrete diffusion VLA backbone and achieves strong performance on both LIBERO and RoboTwin 2.0. Beyond overall success rates, our analyses show that trajectory-level probability modeling is essential: optimizing only the final denoising step fails to capture the executed action-generation process, while multi-step and mixed-step denoising objectives lead to more favorable RL optimization. Furthermore, the proposed Hybrid strategy leverages the flexibility of the trajectory-level formulation to balance task performance and computational efficiency by adapting the denoising steps used during training and inference. These results demonstrate that native discrete diffusion policies, when equipped with principled trajectory-level RL optimization, provide a promising foundation for unified and efficient robotic action generation.

\subsubsection*{Acknowledgments}
We would like to express our sincere gratitude to the Baidu AI Cloud Baige Team for their exceptional technical support and for providing access to the state-of-the-art Baidu AIHC platform. We specifically appreciate the platform's powerful capabilities in delivering efficient training acceleration and enabling ultra-low-latency inference, which were instrumental in optimizing our system's performance during evaluation. The advanced system optimizations and robust distributed infrastructure provided by this team were crucial in accelerating our experiments and validating the scalability of our proposed methods.

Furthermore, we extend our appreciation to the D-Robotics Team for their support with the RoboGo platform. As an end-to-end development and deployment platform tailored for robotic AI, RoboGo significantly streamlined our workflow through its comprehensive toolchain that encompasses model quantization, inference serving, and asset management. These core capabilities facilitated a seamless transition from model optimization to deployment, greatly lowering the technical barriers.

\bibliography{iclr2026_conference}
\bibliographystyle{iclr2026_conference}

\appendix
\clearpage
\section{Pseudocode of UDVLA-RL}
\label{app:pseudocode}


\begin{algorithm}[H]
\caption{Unified Discrete VLA Reinforcement Learning (\udvlarl)}
\label{alg:dvla_rl}
\begin{algorithmic}[1]
\State \textbf{Input:} Pre-trained discrete diffusion VLA model $\pi_\theta$, Value network $V_\phi$, Environment $\mathcal{E}$
\State \textbf{Hyperparameters:} Denoising steps $K$, PPO clip ratio $\epsilon$, PPO epochs $E$, batch size $B$
\State \textbf{Initialize:} Policy parameters $\theta \leftarrow \theta_0$, Value parameters $\phi \leftarrow \phi_0$

\For{iteration $= 1, 2, \dots$}
    \State \textcolor{gray}{\# Phase 1: Environment Rollout \& Trajectory Sampling}
    \State Initialize experience buffer $\mathcal{D} = \emptyset$
    \For{$t = 1 \dots T_{\mathrm{env}}$}
        \State Observe multi-modal state $s_t$ (images, text instructions, proprioception)
        \State Initialize fully masked action sequence $x_K = \texttt{[MASK]}^{N}$
        \For{$k = K$ \textbf{down to} $1$}
            \State Compute confidence scores $p_\theta(\cdot | x_k, s_t)$ via forward pass
            \State Sample update subset $U_k \sim P_{\mathrm{sched}}(U_k | p_\theta, k)$ via Gumbel-TopK
            \State Identify newly decoded tokens subset $M_k \subseteq U_k$
            \State Sample $x_{k-1}^i \sim p_\theta(x_{k-1}^i | x_k, s_t)$ for $i \in U_k$
        \EndFor
        \State Execute final action $x_0$ in $\mathcal{E}$, receive reward $r_t$ and next state $s_{t+1}$
        \State Estimate value $V_\phi(s_t)$ using the first-step hidden state $h^{(0)}$
        \State Store transition $(s_t, x_0, \{x_k\}_{k=1}^K, \{M_k\}_{k=1}^K, r_t)$ in $\mathcal{D}$
    \EndFor
    \State Compute Advantages $A_t$ using GAE over collected rewards and values

    \State \textcolor{gray}{\# Phase 2: PPO Update}
    \For{epoch $= 1 \dots E$}
        \For{mini-batch sampled from $\mathcal{D}$}
            \State \textcolor{gray}{\# Expand batch across $K$ timesteps to compute parallel transitions}
            \State Initialize shielded log-probability $\log \pi_\theta(\tau | s_t) \leftarrow 0$
            
            \For{$k = K$ \textbf{down to} $1$}
                \State Mask target action $x_0$ using inverse of $U_k$ to construct $x_k$
                \State Compute logits $p_\theta(\cdot | x_k, s_t)$ via network forward pass
                \State \textcolor{gray}{\# Drop $P_{\mathrm{sched}}$, only compute grads on $M_k$}
                \State $\log \pi_\theta(\tau | s_t) \mathrel{+}= \sum_{i \in M_k} \log p_\theta(x_0^i | x_k, s_t)$
            \EndFor
            
            \State \textcolor{gray}{\# Calculate PPO Objective}
            \State $r_t(\theta) = \exp \left( \log \pi_\theta(\tau | s_t) - \log \pi_{\theta_{\mathrm{old}}}(\tau | s_t) \right)$
            \State $L^{\mathrm{\udvlarl}}(\theta) = \mathbb{E}_{\tau} \left[ \min\left( r_t(\theta) A_t, \mathrm{clip}(r_t(\theta), 1-\epsilon, 1+\epsilon) A_t \right) \right]$
            \State $L^{\mathrm{VF}}(\phi) = \mathbb{E}_{s_t} \left[ \left( V_\phi(h^{(0)}_{s_t}) - V_t^{\mathrm{target}} \right)^2 \right]$
            
            \State \textcolor{gray}{\# Network Updates}
            \State Update $\theta$ by maximizing $L^{\mathrm{\udvlarl}}(\theta)$
            \State Update $\phi$ by minimizing $L^{\mathrm{VF}}(\phi)$
        \EndFor
    \EndFor
\EndFor
\end{algorithmic}
\end{algorithm}

\clearpage
\section{SFT Backbone and Evaluation Details}
\label{app:sft_details}

\subsection{SFT data and training setup.}
For fair comparison with existing VLA-RL studies, we use the same scale of SFT demonstrations as SimpleVLA-RL. 
On LIBERO, the SFT stage uses 500 demonstrations per suite. 
On RoboTwin 2.0, the SFT stage uses 1,000 demonstrations per task. 
Different from SimpleVLA-RL, which fine-tunes policies for individual tasks or suites, our LIBERO and RoboTwin 2.0 SFT checkpoints are obtained through multi-task joint training. 
Thus, the SFT initialization used by \udvlarl is a unified multi-task discrete diffusion VLA policy rather than a separately fine-tuned task-specific policy.

\subsection{MM-ACT* implementation.}
\textsc{MM-ACT}* follows the MM-ACT architecture and training recipe, with several modifications for efficient RL initialization. 
First, we use action-only SFT, while the original MM-ACT additionally studies multimodal generation objectives. 
Second, we set the action chunk size to 16 for both LIBERO and RoboTwin 2.0, whereas the original MM-ACT uses a chunk size of 8 on LIBERO. 
Third, we use a batch size of 128 on LIBERO and 96 on RoboTwin 2.0. 
Except for the action-only training setting, the action chunk size, and the batch size, all other SFT hyperparameters follow the original MM-ACT setting.

\subsection{Checkpoint selection.}
On LIBERO, the Object suite converges faster than the other suites. 
Therefore, we use the epoch-1 SFT checkpoint for LIBERO-Object and the epoch-4 SFT checkpoint for LIBERO-Spatial, LIBERO-Goal, and LIBERO-Long. 
For RoboTwin 2.0, we use the epoch-4 SFT checkpoint. 
These SFT checkpoints are used as the initialization for subsequent online RL post-training.

\subsection{SFT checkpoint evaluation.}
For LIBERO, we evaluate each subtask with 50 trials, resulting in 2,000 trials over the four LIBERO suites. 
For RoboTwin 2.0, we evaluate the SFT checkpoint under 4-step denoising with 100 trials per task. 
Each evaluation scenario is generated with an independent scenario seed.

\section{LIBERO RL Training Details}
\label{app:libero_rl_training_details}

This appendix summarizes the reinforcement learning training configuration used for the LIBERO experiments. 
For all LIBERO suites, we use the same MM-ACT backbone, action representation, RL algorithm configuration, sampling strategy, and optimization hyperparameters. 
The denoising step number is fixed to 2 for all suites. 
The only suite-dependent setting is the maximum rollout horizon, which is adjusted according to the task horizon of each benchmark suite.

\begin{table}[!htbp]
    \centering
    \caption{Shared LIBERO RL training hyperparameters. These parameters are kept identical across LIBERO-Spatial, LIBERO-Object, LIBERO-Goal, and LIBERO-Long.}
    \label{tab:libero_shared_rl_hyperparams}
    \small
    \begin{tabular}{lc}
        \toprule
        Parameter & Value \\
        \midrule
        Action representation & ee \\
        Action dimension & 7 \\
        Action chunk size & 16 \\
        Denoising steps & 2 \\
        Image resolution & $256\times256$ \\
        Precision & bf16 \\
        Global batch size & 512 \\
        Micro batch size & 8 \\
        Actor learning rate & $3\times10^{-5}$ \\
        Value head learning rate & $2\times10^{-3}$ \\
        Optimizer & AdamW \\
        Adam $\epsilon$ & $1\times10^{-8}$ \\
        Adam $\beta_1,\beta_2$ & 0.9, 0.95 \\
        Weight decay & 0.01 \\
        Gradient clipping norm & 1.0 \\
        Critic warmup steps & 50 \\
        PPO update epoch & 1 \\
        Rollout trajectories per step & 512 \\
        Discount factor $\gamma$ & 0.99 \\
        GAE $\lambda$ & 0.95 \\
        PPO clip ratio & 0.2 \\
        Value clipping & 0.2 \\
        Advantage normalization & Yes \\
        Reward type & chunk-level \\
        Log-probability type & chunk-level \\
        Use wrist camera & Yes \\
        Use proprioception & Yes \\
        \bottomrule
    \end{tabular}
\end{table}

\begin{table}[!htbp]
    \centering
    \caption{Suite-specific rollout horizons for LIBERO RL training.}
    \label{tab:libero_suite_rollout_horizon}
    \small
    \begin{tabular}{lc}
        \toprule
        Suite & Maximum rollout steps \\
        \midrule
        LIBERO-Spatial & 256 \\
        LIBERO-Object & 320 \\
        LIBERO-Goal & 320 \\
        LIBERO-Long & 512 \\
        \bottomrule
    \end{tabular}
\end{table}

As shown in Table~\ref{tab:libero_shared_rl_hyperparams}, the LIBERO experiments use a unified RL training setup across all suites. 
The policy is initialized from the same MM-ACT checkpoint and optimized with the same actor-critic objective, PPO clipping configuration, sampling parameters, and batch settings. 
Table~\ref{tab:libero_suite_rollout_horizon} lists the only suite-specific difference: the maximum rollout horizon. 
Shorter-horizon suites such as LIBERO-Spatial use a rollout limit of 256 steps, while the long-horizon LIBERO-Long suite uses a larger limit of 512 steps to accommodate multi-stage manipulation tasks.

\begin{figure}[!htbp]
    \centering
    \begin{minipage}{0.48\linewidth}
        \centering
        \includegraphics[width=\linewidth]{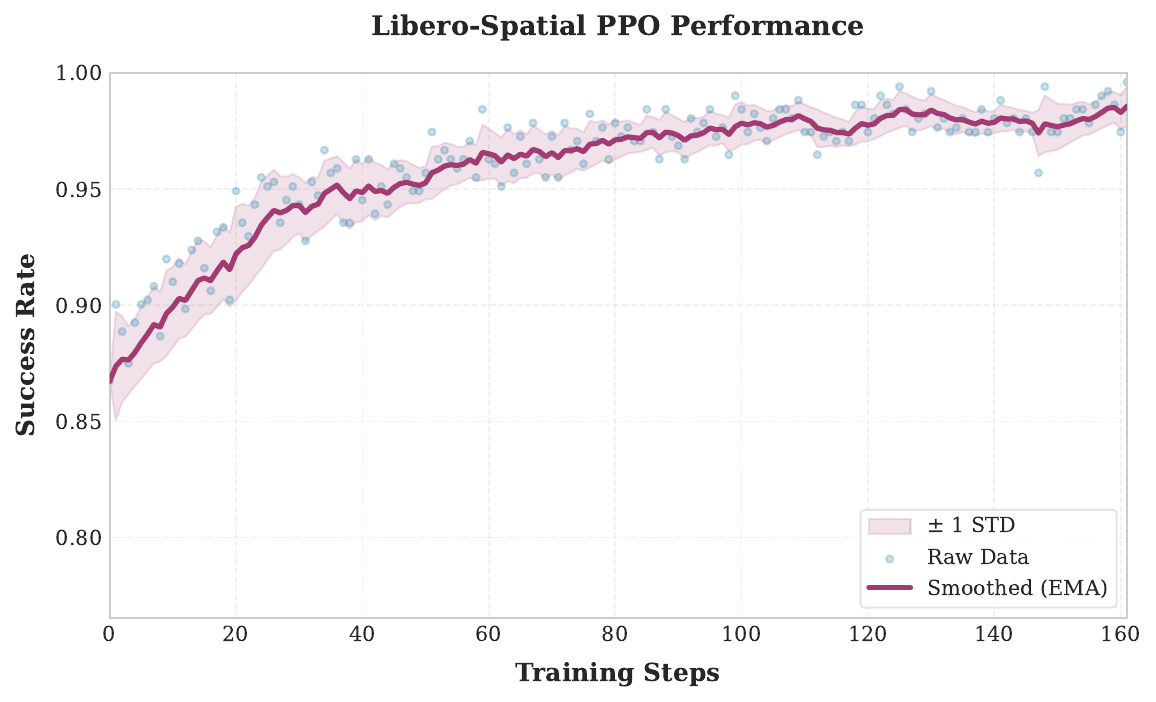}
    \end{minipage}
    \hfill
    \begin{minipage}{0.48\linewidth}
        \centering
        \includegraphics[width=\linewidth]{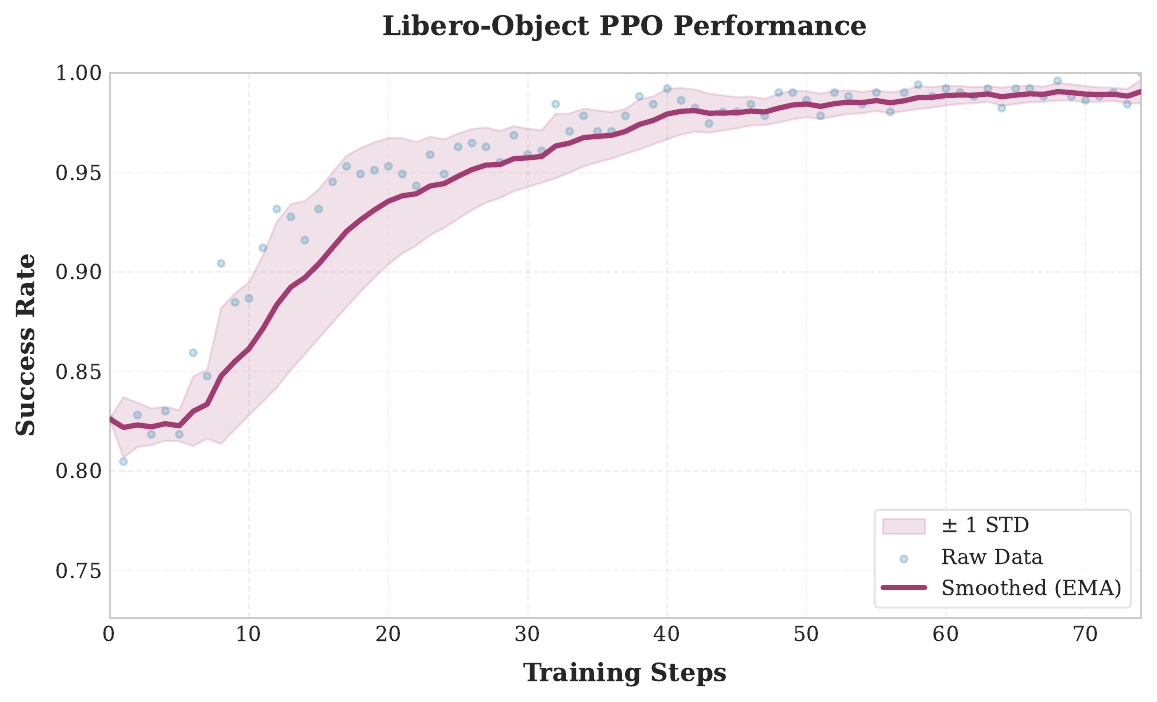}
    \end{minipage}

    \vspace{0.8em}

    \begin{minipage}{0.48\linewidth}
        \centering
        \includegraphics[width=\linewidth]{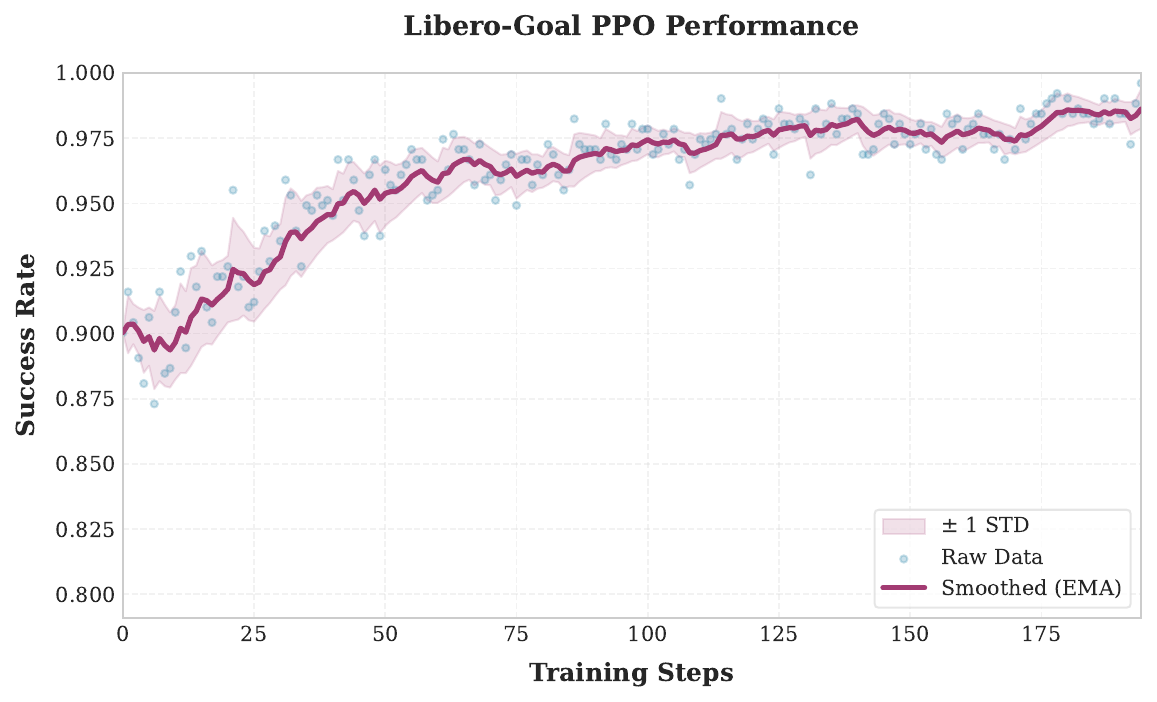}
    \end{minipage}
    \hfill
    \begin{minipage}{0.48\linewidth}
        \centering
        \includegraphics[width=\linewidth]{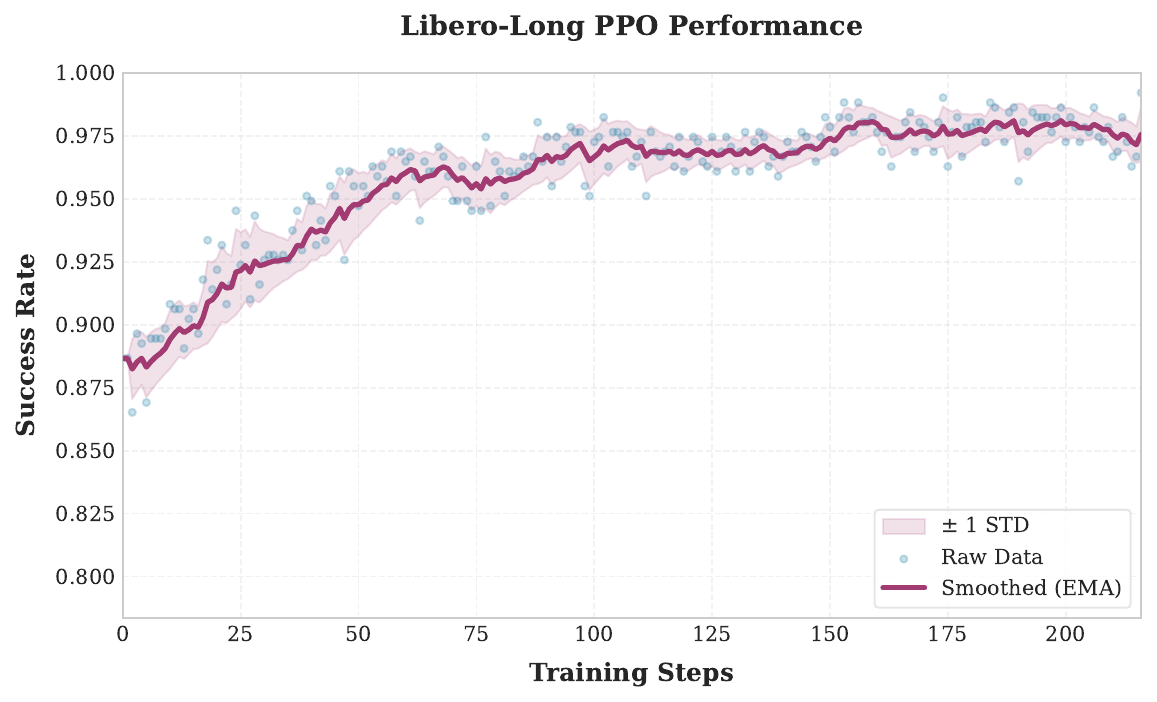}
    \end{minipage}
    \caption{PPO training curves on the LIBERO benchmark suites. Blue points denote raw evaluation results, the smoothed curve denotes the EMA success rate, and the shaded region denotes $\pm 1$ standard deviation.}
    \label{fig:libero_ppo_training_curves}
\end{figure}

Figure~\ref{fig:libero_ppo_training_curves} shows the PPO training curves on the four LIBERO suites. 
Across LIBERO-Spatial, LIBERO-Object, LIBERO-Goal, and LIBERO-Long, the success rate consistently increases during training and gradually converges to a high-performance region. 
The EMA-smoothed curves indicate stable policy improvement despite the variance of raw rollout evaluations. 
Compared with the shorter-horizon suites, LIBERO-Long uses a larger rollout horizon and exhibits a longer training process, which is consistent with the higher complexity of long-horizon multi-stage manipulation tasks.

\section{RoboTwin RL Training Details}
\label{app:robotwin_rl_training_details}

This appendix summarizes the reinforcement learning training configuration used for the RoboTwin experiments. 
To ensure a fair comparison among different denoising-step settings, the 1-step, 2-step, and Hybrid variants use the same training hyperparameters, model configuration, sampling strategy, and environment setup. 
The only difference among these variants lies in the denoising-step configuration: the 1-step and 2-step settings use a fixed number of denoising steps for all tasks, whereas the Hybrid setting assigns task-specific denoising steps according to the initial performance and computational cost of each task.
For online RL training, each RoboTwin task is associated with a pool of 1,000 distinct environment initialization seeds, which instantiate different training scenarios. 
At every RL optimization step, we randomly sample 64 seeds from the task-specific pool and collect one rollout trajectory for each sampled seed. 
Since the training batch contains eight RoboTwin tasks, this produces a total of $64 \times 8 = 512$ rollout trajectories per optimization step. 
The same seed-sampling protocol is used for all denoising-step variants, ensuring that their differences are not caused by different rollout budgets or environment distributions.

\begin{table}[!htbp]
    \centering
    \caption{Shared RoboTwin RL training hyperparameters. These parameters are kept identical for the 1-step, 2-step, and Hybrid settings.}
    \label{tab:robotwin_shared_rl_hyperparams}
    \small
    \begin{tabular}{lc}
        \toprule
        Parameter & Value \\
        \midrule
        Action representation & qpos \\
        Action dimension & 14 \\
        Action chunk size & 16 \\
        Precision & bf16 \\
        Global batch size & 512 \\
        Micro batch size & 8 \\
        Actor learning rate & $3\times10^{-5}$ \\
        Value head learning rate & $2\times10^{-3}$ \\
        Optimizer & AdamW \\
        Adam $\epsilon$ & $1\times10^{-8}$ \\
        Adam $\beta_1,\beta_2$ & 0.9, 0.95 \\
        Weight decay & 0.01 \\
        Gradient clipping norm & 1.0 \\
        PPO update epoch & 1 \\
        Rollout trajectories per step & 512 \\
        Discount factor $\gamma$ & 0.99 \\
        GAE $\lambda$ & 0.95 \\
        PPO clip ratio & 0.2 \\
        Advantage normalization & Yes \\
        Reward type & chunk-level \\
        Log-probability type & chunk-level \\
        Use wrist camera & Yes \\
        Use proprioception & Yes \\
        Maximum episode steps & 256 \\
        \bottomrule
    \end{tabular}
\end{table}

\begin{table}[!htbp]
    \centering
    \caption{Task-specific denoising steps in the RoboTwin Hybrid setting.}
    \label{tab:robotwin_hybrid_task_steps}
    \small
    \begin{tabular}{lc}
        \toprule
        Task & Denoising steps \\
        \midrule
        \texttt{beat\_block\_hammer} & 1 \\
        \texttt{place\_phone\_stand} & 1 \\
        \texttt{place\_empty\_cup} & 1 \\
        \texttt{lift\_pot} & 2 \\
        \texttt{move\_can\_pot} & 2 \\
        \texttt{place\_a2b\_left} & 2 \\
        \texttt{handover\_mic} & 4 \\
        \texttt{pick\_dual\_bottles} & 4 \\
        \bottomrule
    \end{tabular}
\end{table}

For the fixed-step baselines, all RoboTwin tasks use the same denoising-step number during both rollout and evaluation. 
Specifically, the 1-step setting uses one denoising step for every task, and the 2-step setting uses two denoising steps for every task. 
In contrast, the Hybrid setting uses the task-specific denoising-step assignment shown in Table~\ref{tab:robotwin_hybrid_task_steps}. 
This design preserves the same RL training protocol while allowing the denoising budget to better match the difficulty and efficiency requirement of each task.

\subsection{RoboTwin Training-Curve Visualizations}
\label{app:robotwin_training_curves}

\begin{figure}[!htbp]
    \centering
    \begin{subfigure}[t]{0.48\linewidth}
        \vspace{0pt}
        \centering
        \includegraphics[width=\linewidth]{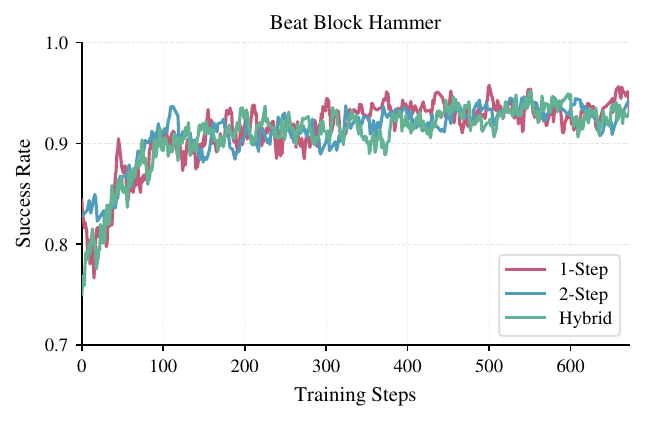}
        \label{fig:robotwin_curve_beat_block_hammer}
    \end{subfigure}
    \hfill
    \begin{subfigure}[t]{0.48\linewidth}
        \vspace{0pt}
        \centering
        \includegraphics[width=\linewidth]{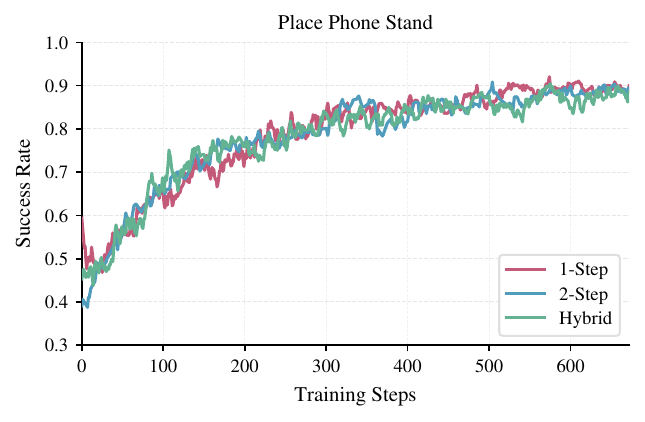}
        \label{fig:robotwin_curve_place_phone_stand}
    \end{subfigure}

    \vspace{0.5em}

    \begin{subfigure}[t]{0.48\linewidth}
        \vspace{0pt}
        \centering
        \includegraphics[width=\linewidth]{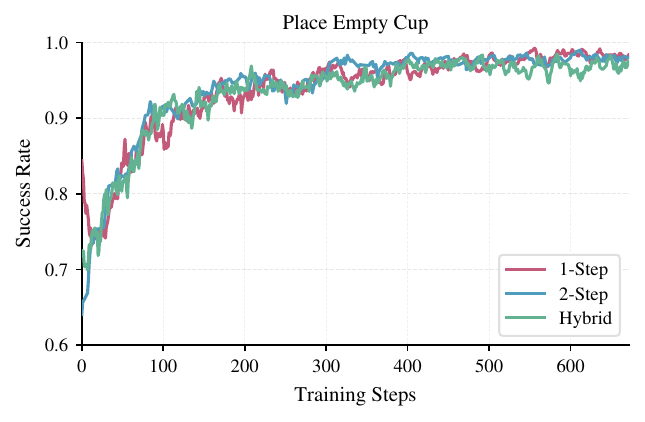}
        \label{fig:robotwin_curve_place_empty_cup}
    \end{subfigure}
    \hfill
    \begin{subfigure}[t]{0.48\linewidth}
        \vspace{0pt}
        \centering
        \includegraphics[width=\linewidth]{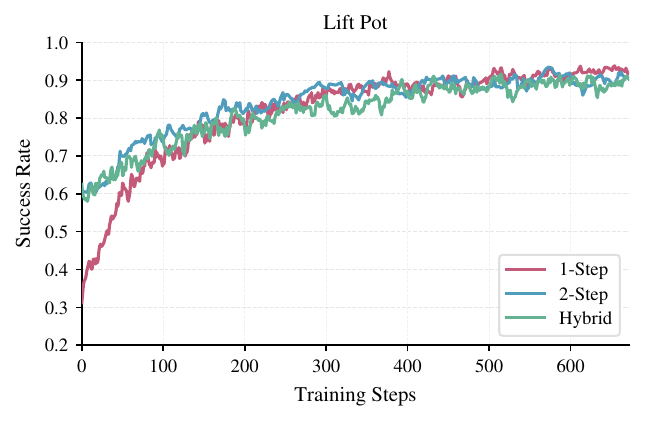}
        \label{fig:robotwin_curve_lift_pot}
    \end{subfigure}
    \caption{RoboTwin training curves on Beat Block Hammer, Place Phone Stand, Place Empty Cup and Lift Pot. The curves compare 1-Step, 2-Step, and Hybrid in terms of success rate over RL training. For all of 8 tasks, training uses a pool of 1,000 distinct environment initialization seeds, from which 64 seeds are randomly sampled for rollout at every RL optimization step.}
    \label{fig:robotwin_training_curves_part1}
\end{figure}

\begin{figure}[!htbp]
    \centering
    \begin{subfigure}[t]{0.48\linewidth}
        \vspace{0pt}
        \centering
        \includegraphics[width=\linewidth]{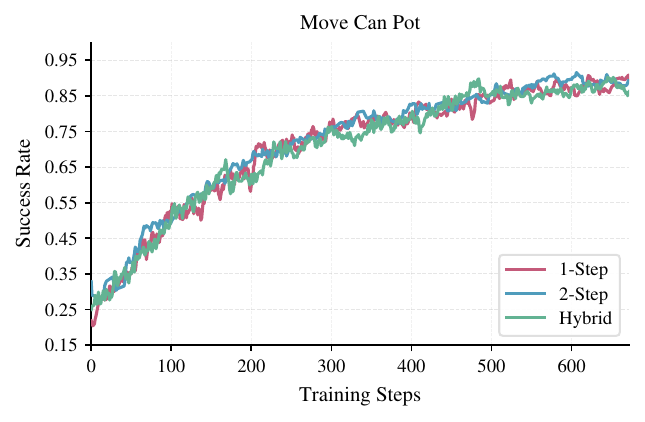}
        \label{fig:robotwin_curve_move_can_pot}
    \end{subfigure}
    \hfill
    \begin{subfigure}[t]{0.48\linewidth}
        \vspace{0pt}
        \centering
        \includegraphics[width=\linewidth]{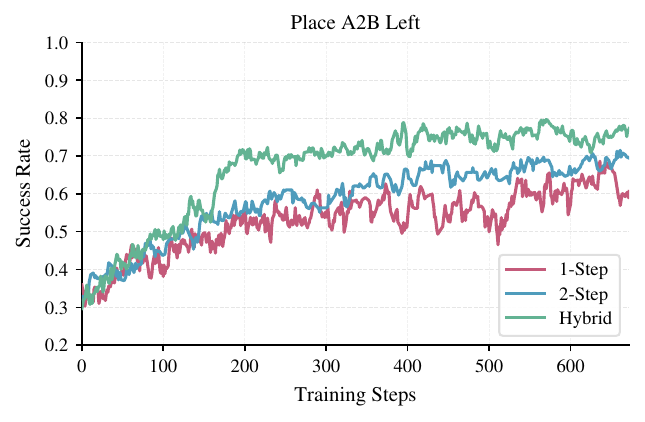}
        \label{fig:robotwin_curve_place_a2b_left}
    \end{subfigure}

    \vspace{0.5em}

    \begin{subfigure}[t]{0.48\linewidth}
        \vspace{0pt}
        \centering
        \includegraphics[width=\linewidth]{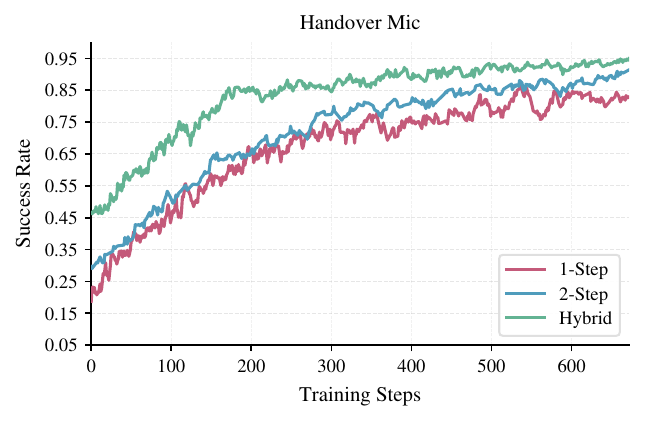}
        \label{fig:robotwin_curve_handover_mic}
    \end{subfigure}
    \hfill
    \begin{subfigure}[t]{0.48\linewidth}
        \vspace{0pt}
        \centering
        \includegraphics[width=\linewidth]{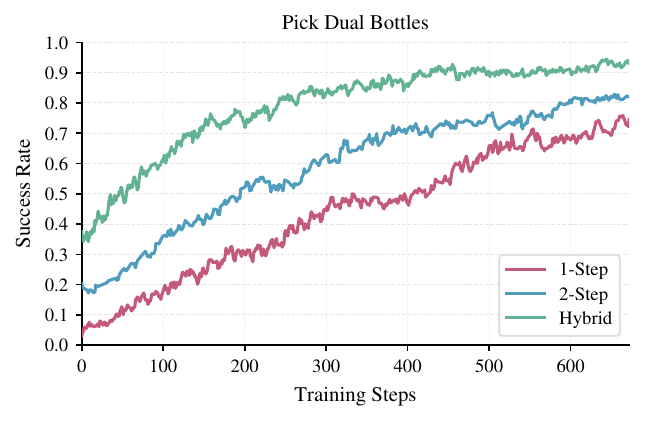}
        \label{fig:robotwin_curve_pick_dual_bottles}
    \end{subfigure}
    \caption{RoboTwin training curves on Move Can Pot, Place A2B Left, Handover Mic, and Pick Dual Bottles, comparing 1-Step, 2-Step, and Hybrid. }
    \label{fig:robotwin_training_curves_part2}
\end{figure}

Figures~\ref{fig:robotwin_training_curves_part1} and~\ref{fig:robotwin_training_curves_part2} provide the complete training-curve visualizations for the eight RoboTwin tasks. Each plot compares the success-rate evolution of 1-step (PD), 2-step, and the task-adaptive Hybrid strategy under the same RL training protocol.

\suppressfloats[t]

\end{document}